\documentclass[10pt,a4paper,conference]{IEEEtran}
\ifCLASSINFOpdf
  \usepackage[pdftex]{graphicx}
  \graphicspath{{../pdf/}{../jpeg/}}
  \DeclareGraphicsExtensions{.pdf,.jpeg,.png}
\else
  \usepackage{graphicx}
  \graphicspath{{../eps/}}
  \DeclareGraphicsExtensions{.eps}
\fi
\usepackage{threeparttable}
\usepackage{hyperref}

\begin{document}
%
\title{Predicting Chemical Properties using Self-Attention Multi-task Learning based on SMILES Representation}

\author{\IEEEauthorblockN{Sangrak Lim and Yong Oh Lee}
\IEEEauthorblockA{Smart Convergence Grop, Korea Institute of Science and Technology Europe Forschungsgesellschaft mbH,\\ Campus E7 1, 66123 Saarbrücken, Germany\\
Email: sangrak.lim@kist-europe.de, yongoh.lee@kist-europe.de}
}


%


\maketitle

\begin{abstract}
In the computational prediction of chemical compound properties, molecular descriptors and fingerprints encoded to low dimensional vectors are used. The selection of proper molecular descriptors and fingerprints is both important and challenging as the performance of such models is highly dependent on descriptors. To overcome this challenge, natural language processing models that utilize simplified molecular input line-entry system as input were studied, and several transformer-variant models achieved superior results when compared with conventional methods. In this study, we explored the structural differences of the transformer-variant model and proposed a new self-attention based model. The representation learning performance of the self-attention module was evaluated in a multi-task learning environment using imbalanced chemical datasets. The experiment results showed that our model achieved competitive outcomes on several benchmark datasets. The source code of our experiment is available at https://github.com/arwhirang/sa-mtl and the dataset is available from the same URL.
\end{abstract}


%
\IEEEpeerreviewmaketitle

\section{Introduction}
Computational methods for predicting the properties of chemical compounds have garnered significant attention in recent decades \cite{qsarModeling, compToxicol, qsarPerspec}. Regulatory agencies and pharmaceutical companies faced a challenge in evaluating the properties of more than 140 million chemicals \cite{chemicalAS}. Traditional \textit{in vivo} and \textit{in vitro} methods have limited abilities in assessing a large number of chemicals, because those methods are neither time nor cost efficient.

Various computational methods, such as read-across \cite{insilicoTox}, dose and time response \cite{foodForT}, toxicokinetics and toxicodynamics, and Quantitative Structure-Activity Relationship (QSAR) models \cite{compToxicol}, have been adopted for chemical compound prediction. In this study, we focus on the QSAR model, which is a statistical-mathematical model used to establish an approximate relationship between a compound's biological property and its structural features \cite{qsarPerspec, qsarWhere, bestPqsar}.

Selecting molecular-structural features and encoding the selected features into fixed-length strings or vectors is a significant challenge for machine learning QSAR models. Conventional molecular machine learning methods implemented sophisticated information to learn the electronic or topological features of molecules from limited amounts of data such as molecular descriptors and fingerprints. Molecular descriptors and fingerprints can be computed by a cheminformatics library such as RDKit \cite{rdkit} using Simplified Molecular Input Line-Entry System(SMILES) \cite{SMILES}, which is a standard approach to represent compounds in the form of strings over fixed characteristics. However, deep learning models based on fingerprints have two problems. First, the SMILES input is transformed into fingerprint; therefore, an additional conversion process is required when interpreting the training results of the model. Second, the search space for certain substructures of chemicals converted into fingerprints can be limited or ignored. In the case of a Morgan fingerprint, different atom environments may be mapped to the same bit.

Inspired by the language model from Natural Language Processing (NLP), several studies have proposed using SMILES as an input for training a set of embedding vectors. Given a natural language sentence, we can represent a word or a character with embedding vectors and train deep learning models based on those vectors \cite{word2vec}. Cadeddu et al. \cite{nlpIntoBio} found that a chemical compound has a similar structure to a natural language sentence in terms of frequency through quantitative analysis. If a chemical compound corresponds to a sentence, then each character notation in SMILES corresponds to a character in this regard. Several attempts to use SMILES as input for predicting certain chemical properties have been made because the embedding vectors can represent semantically similar items because these items are located near the vector spaces \cite{SCFP,fp2vec,smiles2vec,moleculenet}. Once a machine learning model has been trained on a corpus, the embedding vectors, as well as the trained model, are applicable to various datasets. Compared to using fingerprints for a machine learning model, the embedding vectors can be fine-tuned by datasets, whereas fingerprints are fixed after they are generated.

The transformer model and its variants have exhibited state-of-the-art performance in various datasets since its introduction in 2017 \cite{self_attention}. The transformer model implemented a concept of multi-head self-attention. The self-attention method allows the extraction of different information from a sentence for classification datasets \cite{self_attention2}. The transformer model can learn the long-range dependencies in the input data, and the model does not include recurrent connections; thus paralleled computation is feasible. Owing to the interest garnered by the transformer model in the field of chemical compound property classification, we have found several studies related to the self-attention model within the aforementioned field \cite{middleProduct,manyDataT,SA_centered}. However, we also believe that improvements should be made to the existing models. Thus, we proposed a novel self-attention based deep learning QSAR model that uses SMILES as input without computing descriptors and fingerprints. This study will focus on the differences between our model and other existing models that have also applied self-attention because other studies have examined similar structures.

The proposed method also included multi-task learning, which can be implemented with shared hidden layers and discrete output layers \cite{MTL}. The output layers are separately trained only for a specific task. Some of the beneficial traits of multi-task learning include generalization and implicit augmentation of training samples. Training multiple tasks reduces the risk of overfitting because various tasks have different amounts of noise. Furthermore, the multi-task structure forces the shared hidden layers to focus on important and inherent features that are common in the training samples of several tasks. Deep learning models have a feature known as representation learning that automatically extracts useful information, and multi-task learning is such an example because its shared hidden layers capture an underlying subset of information that may be relevant for each particular task \cite{representLearn}. To create a synergy from different tasks, both the input data and given tasks should resemble each other throughout the tasks. Although defining multiple task resemblance is not easy \cite{mtl_dp}, chemical compound property prediction generally shares several features between tasks.

We propose Self-attention Multi-Task learning (SA-MTL) as a QSAR model. The contributions of our study as follows:
\begin{itemize}
  \item Our method achieved state-of-the-art performance in several datasets.
  \item We describe the structural difference of each transformer -variant model, and show the influence of such a structural change on learning.
  \item We introduce implementation details of the smiles embedding for analyzing the trained model.

\end{itemize}

In the following sections, we will explain our methods and describe our experiments. Five subsections explain independent datasets in the experiments. Finally, we discuss our conclusions in the last section.

\section{Related Work}
Wu et al.\cite{moleculenet} claimed that creating a new benchmark for every new article prevented the development of chemical compound property prediction, owing to the absence of standard benchmark. Even if some studies evaluated the same data, the post-processing or split method might differ. Wu et al. set up a website with several downloadable datasets and presented them together with the results of previously proposed machine learning models. The HIV, SIDER, BBBP, and clintox dataset from the website\footnote{http://moleculenet.ai/} are used for our experiment. The models ranging from conventional machine learning to deep learning that have shown the best performance in each dataset from the website \cite{moleculenet} are compared to our proposed method. 

In the SMILES Convolution FingerPrint (SCFP) model \cite{SCFP}, each SMILES sequence for a chemical compound is represented with 21-bit atomic characteristics and 21-bit other chemical characteristics to compute the SMILES feature matrix. In the FingerPrint To VECtors (FP2VEC) model \cite{fp2vec}, the selected extended-connectivity fingerprints \cite{ECFP} of chemical compounds are embedded within the matched randomized vector in the lookup table. The FP2VEC \cite{fp2vec} and SCFP \cite{SCFP} models are based on similar principles. These deep learning models can replace the molecular fingerprint, and a small number of Convolution Neural Network(CNN) layers can provide sufficient learning capability. The SCFP model proposed a new method of adding an embedding layer to SMILES by using the custom word embedding method. The authors of the SCFP model focused only on Tox21 data, and they provided pre-processing data that eliminated redundant or problematic instances of Tox21 data; therefore, we used their version of data for the Tox21 task \cite{TOX21}. The essential feature of the FP2VEC model was the multi-task learning. The authors of the FP2VEC model showed that multi-task learning is a practical approach in the compound property prediction domain.


\begin{figure*}[!t]
\includegraphics[scale=0.5]{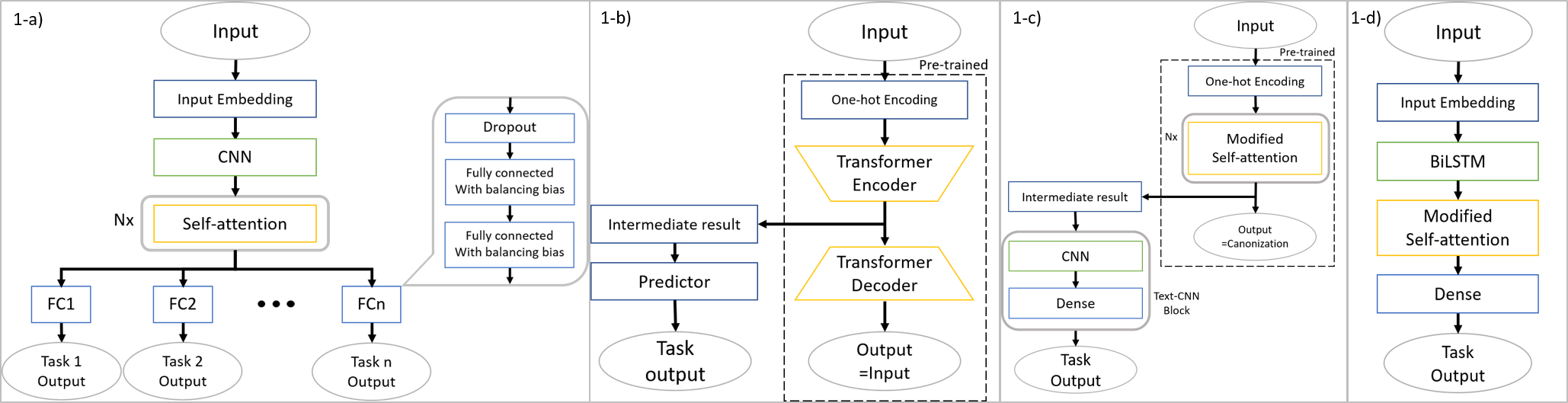}
\centering
\caption{{\bf Model architectures of SA-MTL and existing self-attention based approaches.}
Simplified overview of our model and existing models. For comparison, the three existing models are depicted with core components only. \textbf{1-a) SA-MTL Model}: Our SA-MTL consists of three deep learning components. After input embedding, a CNN layer learns the shared underlying factors. The next component is a self-attention module. This self-attention module consists of the encoder part of the transformer model. The last component is the discrete output layer. We represent discrete output layers with FC since each output layer consists of two fully connected layers. The discrete output layers reduce the dimension to match the target tasks. \textbf{1-b) Smiles\_Transformer Model}: The Smiles\_Transformer model uses the intermediate result obtained from the pre-training step. \textbf{1-c) Transformer-CNN Model}: The Transformer-CNN model also implemented the pre-training approach. The model contains text-CNN block for several CNN layers. \textbf{1-d) BiLSTM-SA Model}: The concept of the BiLSTM-SA model implemented a self-attention module without the multi-task learning scheme. All the three existing studies chose a dense layer as a final predictor, while we adopted the discrete output layer.}
\label{fig1}
\end{figure*}

We introduce following Transformer-variant models or self-attention using models.

The Smiles\_Transformer model \cite{middleProduct} devised a way to utilize both the encoder and decoder structures of the original transformer model. The authors attempted to implement a pretraining approach that was similar to the original transformer model. During the pretraining phase, the authors provided SMILES representation of a compound for the encoder model and then set the target of the decoder as a different SMILES of the same compound. To generate the target SMILES, they used the SMILES-Enumerator. After unsupervised pretraining, the Smiles\_Transformer model used the hidden layer generated by the encoder as a molecular fingerprint and applied a fully connected layer for any tasks to predict results. 

The Transformer-CNN model \cite{manyDataT} utilized a similar pretraining method as to the Smiles\_Transformer model. The input used for pretraining is the generated SMILES of a compound, and the output is a canonical SMILES of the compound; hence the process is very similar to the Smiles\_Transformer method described above. Whereas the Smiles\_Transformer uses both the encoder and decoder of the original transformer model, Transformer-CNN model uses only the encoder. The authors claimed that the hidden layer output of the encoder model constitutes the molecular fingerprint. After the pretraining phase, the Transformer-CNN model implemented several layers of CNN. The location of the CNN layers in the Transformer-CNN is different from that of our SA-MTL model. Intuitively, a self-attention module can learn long-range dependency, whereas the maximum range of dependency a CNN layer can learn is defined by a relatively small filter size. We placed our CNN layer before the self-attention module. Thus, the CNN layer mainly performs small-unit learning, and the self-attention module manages the long-range relationships within a chemical compound.

The BiLSTM self-attention (BiLSTM-SA) model \cite{SA_centered} and our SA-MTL model share the same basic concept when it comes to a single task; the first part of the BiLSTM-SA model is BiLSTM, which is a Recurrent Neural Network (RNN) variant; the second part is a modified self-attention module, and; the last part is a dense layer. It is better to use the attention module in the prediction rather than utilizing the attention module in the pre-training phase because the attention module is known for its capability of analyzing a trained model. All the three existing studies chose a dense layer as a final predictor, whereas we adopted the discrete output layer.

\section{Method}
\subsection{Multi-Task Learning Self-Attention Method}

We present a Self-Attention Multi-Task Learning (SA-MTL) model for compound property classification that has three deep learning components. Our entire model is shown in Fig. 1-a). The basic approach for processing SMILES is to regard each symbol as one character and perform character-level embedding because SMILES is a combination of symbols representing atoms and bonds. However, several atoms consist of two characters. When a particular atom is represented by two characters, embedding the atom one by one causes an analyzing error in the corresponding atom part. For instance, when analyzing model results after training the model, the weights of the model must be converted into the same dimension as the input. If the atom of two characters such as "Cl" is simply divided into "C" and "l," the atom "Cl" cannot be converted correctly. Character embedding assigns an index number corresponding to one embedding vector to each letter. We assigned an atom-embedding vector index to an atom consisting of two letters. The "H" atom is usually omitted from SMILES but is expressed in SMILES in some cases. Later, in the analysis phase, it is essential whether a specific letter is an atom or a bond signal. We consider "H" atoms as a bond signal such as "@." It is also important to save the embedding vectors for the analyzation process.
 
The first component is the CNN layer, which serves as a shared hidden layer. It is a two-dimensional convolution with a filter size of 7 with the same padding. According to the FP2VEC model \cite{fp2vec}, even one CNN layer could achieve competitive results. The output shape of the CNN layer is [batch size, sequence size, hidden size]. Adding this CNN layer to the model helps to learn features by sharing multiple tasks in a multi-task learning environment. We considered the following differences in applying the CNN layer to our model; the size of the training dataset is comparatively small and class-imbalanced, and the classification task delivers insubstantial feedback compared with tasks that have a target sentence. In the ablation study, we will discuss the representation learning power of our components.

The second component is the self-attention module. We included the encoder component of the transformer model \cite{self_attention} only, because the decoder component, which generates the target sentence, is not required for the classification. We also found that there was little to none performance degradation when we excluded position embedding from the model. The sinusoidal position embedding was an essential component of the original transformer model that was used to consider the relationship between each word in the source sentence and the corresponding word in the target sentence. However, the influence of the position embedding is limited because the target in our model is a binary class. This also applies to the multi-head structure of the self-attention module. The multi-head code divides the sequence of the SMILES by the number of heads. Each head in the original transformer should attend to information from different representations at different positions.

The final component is the discrete output layer, which consists of two fully connected layers. The two fully connected layers do not change depending on whether the setting is multi-task or single-task. Each of the discrete output layers fine-tunes exclusively based on each of the tasks. The output shape of the self-attention module is [batch size, sequence size, hidden size]. We set the output size for both fully connected layers to 1, because the target classes have a shape of [batch size]. There are other dimension reduction methods such as max-pooling, but we have experimentally demonstrated that SA-MTL structure is more effective. The max-pooling method itself has the effect of reducing the computational load because it does not require any computation of variables. The pooling helps to make the representation invariant to small translations of the input \cite{Deeplearing_ian}. However, in the ablation study, we showed that it could be more effective to implement the fully connected layer twice even in the dimension reduction step than in max-pooling. A balancing bias is applied to the fully connected layers to rectify the class-imbalance in the data. The balancing bias is the ratio of negative to positive instances in the training data.

The weighted cross-entropy function is utilized to calculate the cost function. To alleviate the class imbalance problem, we applied the balancing bias as a weight for the discrete output layer.

\subsection{Difference between our model and the Smiles\_Transformer}
The simplified architecture of the Smiles\_Transformer is shown in Fig. 1-b). The authors of the Smiles\_Transformer used the pre-train approach to learn the atom-level embedding from large unlabeled dataset using the transformer model. Subsequently, they used a simple predictor model based on the intermediate result of the pre-train approach to fine-tune on the datasets. Pre-training approach should be used with caution. In the Tox21 dataset, a specific chemical(“CCC(=O)Nc1ccc(Cl)c(Cl)c1”) in NR-AhR target task is toxic, whereas the same chemical is non-toxic in the NR-ER-LBD target task. When a specific chemical acts as a ligand, the toxicity of the chemical varies depending on the protein target. Therefore, protein targets should be considered, if one would implement pre-training approach on SMILES data.  

\subsection{Difference between our model and the Transformer-CNN}
The simplified architecture of the Transformer-CNN is shown in Fig. 1-c). Similar to the above model, the Transformer-CNN model utilized pre-training approach. The authors of the Transformer-CNN model mentioned about the two-character atoms. Although they claim that dealing with the two-character atoms does not have a very influential performance, the transformer-CNN model achieved higher scores compared to the SMILES\_Transformer model because they used a more complex predictor. We also indicate that the Transformer-CNN model does not implement multi-task learning scheme.  

\subsection{Difference between our model and the BiLSTM-SA}
The simplified architecture of the BiLSTM-SA is shown in Fig. 1-d). The first component of our SA-MTL is a CNN layer, whereas the BiLSTM-SA model uses BiLSTM as the first component. In the ablation study section, we experimented with our model by replacing the CNN component with RNN, and our model showed no significant difference in performance. However, the recurrent structure is inferior to the CNN structure in computational speed. The second difference is that the self-attention module of BiLSTM-SA is not repeated. In the original model, self-attention has a multi-layered structure, which indicates that the multi-layered self-attention is more effective. We chose seven-layered self-attention for our model after hyperparameter optimization.

The main reason for the performance difference is that the BiLSTM-SA model did not implement multi-task learning scheme. All the existing transformer-variant models use a dense or fully connected layer as a final prediction layer. Considering the simple final layers of the existing models, we also appreciate our discrete output layer. As a reminder, the authors of the BiLSTM-SA model did not mention the two-character atoms. The two-character atoms are essential in analyzing the weights of the trained model. However, it is easy to disregard the two-character because not considering the two-character atoms has no significant influence on performance, as claimed by the authors of the Transformer-CNN.

\section{Experiments}
SA-MTL, the proposed method, was evaluated in Tox21, HIV, SIDER, BBBP, and CLINTOX. In the datasets, the chemical compounds were represented as SMILES. The statistics of the number of classes, average instances and positive-negative ratio are presented in Table~\ref{tab1}. Notably, positive-negative ratio is significantly imbalanced in Tox21, HIV, and CLINTOX.

The dataset was divided into training, validation, and test sets. SA-MTL was optimized by selecting the model hyperparameters that maximized the Area Under the Curve(AUC) on validation sets. Table ~\ref{tab2} summarizes the hyperparameter search range, and the selected optimal parameters using random search. The model weights of the best model in each validation set were saved and we measured the prediction performance using the saved model on the test sets. We reported an average of five times of repeated evaluations to achieve more rigorous results. The performance metric was AUC of the Receiver Operating Characteristic curve(ROC-AUC). 

SA-MTL was compared with CNN-based models (SCFP \cite{SCFP} and FP2VEC \cite{fp2vec}), transformer-variant models (BiLSTM-SA \cite{SA_centered}, Transformer\_CNN \cite{manyDataT}, Smiles\_Transformer \cite{middleProduct}), and the other methods in DeepTox \cite{deeptox} along with several excellent models from Wu et al.\cite{moleculenet}. The results except that of our model were taken from the references mentioned above. SA-MTL was implemented using TensorFlow version 2.1. The experiments were conducted in Nvidia Titan RTX GPU. 




\begin{table}
\begin{threeparttable}
\caption{Dataset Statistics}\label{tab1}
\begin{tabular}{lllp{2.1cm}}
\hline
Dataset & Num of Classes & Ave. Instances & Pos/Neg ratio\tnote{\dag}\\
\hline
Tox21 & 12 & 7831 & 1:13.4 \\
HIV & 1 & 41127 & 1:27.4 \\
SIDER & 27 & 1427 & 1:0.75\\
BBBP & 1 & 2031 & 1:0.25\\
CLINTOX & 2 & 1478 & 1:0.06$|$1:12.25\tnote{*}\\
\hline
\end{tabular}
\begin{tablenotes}\footnotesize
\item[\dag] The positive to negative ratio is the total sum value of the training data if the number of classes is more than one except CLINTOX.
\item[*] The CLINTOX has two classes that have different Pos/Neg ratio.
\end{tablenotes}
\end{threeparttable}
\end{table}

%

\begin{table}[!t]
\begin{threeparttable}
\centering
\caption{Search range of finding the optimized hyperparameters}\label{tab2}
\begin{tabular}{llll}
\hline
&Parameter & Test Range & Selected\\
\hline
Common & Hidden Unit Size 1\tnote{\dag} & 50 - 200 & 128 \\
 & Hidden Unit Size 2 & 128 - 2048 & 1024 \\
 &  (used inside of SA\tnote{*} Layer) &  &  \\
 & Embedding Size & 32 - 1024 & 128 \\
 & Filter Size (CNN) & 3 - 20 & 7 \\
 & Batch Size & 32 - 256 & 60 \\ 
 & Learning Rate & 0.00001 - 0.001 & 0.00005 \\
 & Dropout Rate & 0 - 0.5 & 0.1 \\
 & Num of SA layers & 1 - 10 & 5 \\
 & Multi-head & 1 - 8 & 1 \\
\hline
\end{tabular}
\begin{tablenotes}\footnotesize
\item[\dag] The hidden unit size 1 is used for the CNN output hidden size and general hidden size of the self-attention module.
\item[*] SA denotes the self-attention. SA module has a fully connected layer inside . The hidden unit size 2 is used at the fully connected layer.
\end{tablenotes}
\end{threeparttable}
\end{table}

\subsection{Dataset 1: Tox21}
The Tox21 dataset was introduced at the TOX 21 Challenge 2014 \cite{TOX21}. The goal of the challenge was to elucidate the interference of biochemical pathways by compounds that can be inferred from chemical structure data. The dataset contains approximately 8000 compounds of SMILES, which can be classified into 12 target classes. A compound instance can be tagged as multiple classes which makes this data suitable for multi-task learning. 

The Tox21 Challenge presented datasets named "train", "test", and "score". We performed a two-phase evaluation. The first evaluation was performed on the "train" and "test" using the random split method. As observed in Table ~\ref{tab3}, SA-MTL achieved the highest AUC, whereas other transformer-variant models exhibited lower AUC compared with the CNN models. Our SA-MTL model exhibited 0.07 higher AUC than that of the Graph convolution (GC), which is an extensively used deep learning model in toxicity prediction. Otherwise, both the Transformer\_CNN and Smiles\_Transformer achieved a lower performance compared with the GC. For the Transformer\_CNN model, placing the CNN behind the encoder component was not an appropriate option, and the Smiles\_Transformer model only used the transformer model for pretraining. 

The second evaluation was performed on the "score" data after training on other datasets. We found that the score data have a slightly different distribution compared with other data. As presented in Table~\ref{tab3}, SA-MTL without an ensemble achieved a substandard performance compared with the DeepTox model \cite{deeptox} and SCFP \cite{SCFP}. DeepTox, the winning model of the TOX 21 Challenge 2014, used an ensemble technique. For fair comparison, we applied the ensemble accordingly. Our ensemble method summed the five output probabilities of the same model where initial weights were determined differently. The SA-MTL with an ensemble achieved a better performance in "score" data compared with DeepTox and SCFP with an increment of 0.036 AUC to our SA-MTL without an ensemble. The DeepTox model is an ensemble of several different models, and such a model is challenging to re-implement. Meanwhile, our SA-MTL ensemble version uses the multiple results of one SA-MTL model.





\begin{table}
\begin{threeparttable}
\caption{Tox21 evaluation results compared to other models}\label{tab3}
\begin{tabular}{lll}
\hline
 \multicolumn{3}{c}{Comparison results on Train and Test Data} \\
\hline
Model &  Notes & Average AUC \\
\hline
SA-MTL(OURS) & random split & \textbf{0.9}\\
SCFP & cross-validation  & 0.877\\
FP2VEC & random split & 0.876\\
BiLSTM-SA & stratified random split & 0.842 \\
GC\tnote{*}  & random split & 0.829 \\
Transformer\_CNN & cross-validation \& augmented & 0.82 \\
Smiles\_Transformer & random split & 0.802 \\
\hline
 \multicolumn{3}{c}{Comparison results on Score Data} \\
\hline
Model &  Notes & Average AUC \\
\hline
SA-MTL(OURS) & without ensemble & 0.806\\
SA-MTL(OURS) & with ensemble & \bf{0.842}\\
DeepTox\cite{deeptox}\tnote{**} & with ensemble & 0.837\\
SCFP & without ensemble & 0.813\\
\hline
\end{tabular}
\begin{tablenotes}\footnotesize
\item[*] Result from Wu et al.\cite{moleculenet} Original model was introduced by Altae-Tran et al.\cite{graphConv}
\item[**] Result from Mayr et al.\cite{deeptox}\\
Note: The best results on the test set are highlighted in bold.
\end{tablenotes}
\end{threeparttable}
\end{table}

\subsection{Dataset 2: SIDER }
The SIDER dataset contains drugs and their associated adverse reactions that were recorded during clinical trials \cite{sider}. The information was extracted from public documents and package inserts. The dataset has a total of 1427 instances for 27 targets. The organizers of the SIDER dataset also used NLP to extract the adverse drug reactions from package inserts. We used the random split method for the SIDER dataset. The SIDER dataset is similar to the TOX21 dataset in that a compound instance can be tagged as multiple classes. 

The evaluation results are presented in Table~\ref{tab4}. Other results except ours are directly obtained from the papers. Our model exhibited good performance despite fewer data instances and more classes compared with the Tox21 data.

\begin{table}
\begin{threeparttable}
\caption{SIDER evaluation results compared to other models}\label{tab4}
\begin{tabular}{p{2.2cm}p{3.2cm}p{2cm}}
\hline
Model &  Notes & Average AUC \\
\hline
SA-MTL(OURS) & random split & \bf{0.858}\\
FP2VEC & random split & 0.836 \\
RF\tnote{*} & random split & 0.684 \\
\hline
\end{tabular}
\begin{tablenotes}\footnotesize
\item[*] Result from Wu et al. (2018).\\
Note: The best results on the test set are highlighted in bold.
\end{tablenotes}
\end{threeparttable}
\end{table}

\subsection{Dataset 3: HIV}
The HIV dataset was introduced by the NCI Development Therapeutics Program (DTP) AIDS Antiviral Screen, which tested the anti-HIV activity of over 40,000 compounds. The original data has three categories, the binary target class, which Wu et al.\cite{moleculenet} merged the categories into, is used in this evaluation. Most cancers progress through a complex mechanism. This dataset has performance limitations in that only the SMILES of the ligand is provided as an input instead of considering various mechanisms. As recommended \cite{moleculenet}, we used scaffold splitting that separates structurally different molecules into different training or test subset.

The evaluation results are presented in Table~\ref{tab5}. SA-MTL exhibited a reduced AUC compared with the Transformer\_CNN. We separated the results obtained using scaffold split data from other results because scaffold splitting makes the prediction more difficult. In comparison with other models using scaffold split, our SA-MTL achieved better AUC compared to the previous best performing model (KernelSVM) \cite{moleculenet}. For this dataset, a conventional machine learning model achieved a better result compared with the deep learning models. 

\begin{table}
\begin{threeparttable}
\caption{HIV evaluation results compared to other models}\label{tab5}
\begin{tabular}{lll}
\hline
Model &  Notes & Average AUC \\
\hline
Transformer\_CNN & cross-validation \& augmented & \bf{0.83} \\
BiLSTM-SA & stratified random split & 0.8 \\
\hline
 \multicolumn{3}{c}{Evaluation Results with Scaffold Split} \\
\hline
SA-MTL(OURS) & scaffold split & \bf{0.826}\\
KernelSVM\tnote{*} & scaffold split & 0.792 \\
FP2VEC & scaffold split & 0.754 \\
Smiles\_Transformer & scaffold split & 0.729\\
\hline
\end{tabular}
\begin{tablenotes}\footnotesize
\item[*] Result from Wu et al.\cite{moleculenet}.\\
Note: The best results on the test set are highlighted in bold.
\end{tablenotes}
\end{threeparttable}
\end{table}

\subsection{Dataset 4\&5: BBBP \& clintox}
The human blood-brain barrier is a membrane barrier that protects the nervous system. If a new drug targets the nervous system, the blood-brain barrier should be considered. The authors of the BBBP dataset compiled the data from a number of publications regarding the human blood-brain barrier \cite{bbbp}. The initial dataset contains a total of 2052 compounds, but some chemical compounds are filtered out, because the valence of a certain atom is abnormal. After filtering out, 2031 compounds remained. We applied scaffold split method to evaluate SA-MTL for testing under difficult conditions. However, SA-MTL could achieve AUC score of 0.966. One of the reasons for the high score is the positive to negative ratio. The positive-to-negative ratio of this BBBP dataset is different from the other datasets. 

The CLINTOX dataset was introduced by Wu et al.\cite{moleculenet}. They compiled this dataset using a list of FDA-approved drugs, and a list of drugs that are known to have toxicity. Our model scored over 0.95 AUC on BBBP and Clintox datasets, indicating that both BBBP and Clintox datasets contain no significant noise and the objective of the datasets are suitable for machine learning prediction.

The evaluation results are presented in Table~\ref{tab6}. Other results except ours are directly obtained from the papers.

\begin{table}
\begin{threeparttable}
\caption{BBBP and CLINTOX evaluation results compared to other models}\label{tab6}
\begin{tabular}{llll}
\hline
Dataset & Model &  Notes & Average AUC \\
\hline
BBBP &SA-MTL(OURS) & scaffold split & \textbf{0.954}\\
&SA-MTL(OURS) & random split & 0.945\\
&Transformer\_CNN & CV \& augmented & 0.92 \\
&KernelSVM\tnote{*} & scaffold split & 0.729 \\
&FP2VEC & random split & 0.713 \\
&Smiles\_Transformer & scaffold split & 0.704\\
\hline
CLINTOX &SA-MTL(OURS) & random split & \textbf{0.992}\\
&SA-MTL(OURS) & scaffold split & 0.99\\
&Smiles\_Transformer & scaffold split & 0.954\\
&Weave\tnote{*} & random split & 0.832 \\
&Transformer\_CNN & CV  \& augmented & 0.77 \\
\hline
\end{tabular}
\begin{tablenotes}\footnotesize
\item[*] Result from \cite{moleculenet}. Original model was introduced by Kearnes et al.\cite{weave}\\
Note: The best results on the test set are highlighted in bold.
\end{tablenotes}
\end{threeparttable}
\end{table}

\subsection{Ablation Study}
An ablation study was performed to evaluate the effectiveness of several features in SA-MTL. We started with our best performing model and removed features individually to track any changes in performance. The results was an average of five times testing of models on the Tox21 dataset. \newline

\noindent\textbf{Effect of two-character embedding, self-attention, multi-task learning, and a CNN layer:}\space\space\space\space\space The first of the ablation study is to assign character embedding even for the atoms consisting of two characters. The performance of the character embedding model reaches 0.9 on the validation set. However, we could observe performance drops to 0.87 on test set from time to time; we attribute the occasional degradation owing to the random distribution of the two-character atoms. The self-attention module and the multi-task learning scheme are two essential components of our model. First, we evaluated our model without the multi-task learning, and the AUC score was worse than the original SA-MTL model. Second, we excluded self-attention module from our model and the AUC score was 0.798. This result is counter-intuitive because our SA-MTL model without the self-attention module is similar to the FP2VEC model, which scored higher AUC than the current model. After several experiments, we observed that if we also replace our discrete output layer with a max-pooling layer, the performance improves. For a simple model with an imbalanced data, a max pooling layer can perform better than the discrete output layer. The choice of a layer with potential for information loss during the dimension reduction step has a significant impact on complex components such as self-attention modules.
  
After we conducted an ablation study of the second and the third component of our model, we evaluated our model without the first CNN layer, and the result was worse than the original result. In our multi-task learning environment, the self-attention module was insufficient for learning the shared factors of multiple tasks with imbalanced data. Additional data and more explicit feedback from the objective function would improve the performance. \newline

\noindent\textbf{Effect of the CNN layer and the discrete output layer:}\space\space\space We replaced the first CNN layer with an RNN layer and still achieved appropriate results. The RNN layer was a gated-recurrent unit \cite{GRU}. However, we chose the CNN structure, because the recurrent structure was slower. We also evaluated the final layer by replacing the discrete output layer with a max pooling layer. Our discrete output layer performed better than a max-pooling layer. Again, a max-pooling layer might achieve better results with simple models. \newline

\noindent\textbf{Effect of the multi-head feature and the position encoding:}\space\space\space Although this is an ablation study, we experimented SA-MTL model with more features than the original. We tested five multi-heads and position encoding. We found that the multi-head feature did not have a significant impact on chemical compound prediction. The multi-head feature was devised to attend to information from multiple different positions. For reasons similar to position embedding, the multi-head feature was not very effective for classification. We regard this problem as an over-parametrization issue. Kovaleva et al.\cite{understandBert} showed that most self-attention heads contain trivial information or the heads sometimes attend to the same location. They claim that such redundancy is an example of the over-parametrization.
  
If we add the position embedding value to each atom, the performance of the model remains the same. A SMILES string is different from natural language sentence because the position of an atom does not convey grammatical meanings. When applied to a classification dataset without feedback such as target sentences, position embedding has limited effects on performance.


The ablation study results are presented in Table~\ref{tab7}.  

\begin{table}
\begin{threeparttable}
\caption{Performance changes by modifying several features of our model in the Tox21 dataset.}\label{tab7}
\begin{tabular}{p{1.1cm}p{4.7cm}p{1.7cm}}
\hline
&Modified Features&Average AUC\\
\hline
SA-MTL & & 0.9\\
SA-MTL& - Two-Character Embedding & 0.888\\
SA-MTL& - Multi-task Learning & 0.871\\
SA-MTL& - Self Attention Module  & 0.798\\
SA-MTL& - CNN  & 0.824\\
\hline
SA-MTL& CNN$<>$RNN\tnote{*}  & 0.895\\
SA-MTL& Discrete Output Layer$<>$Max Pooling\tnote{**}  & 0.865\\
\hline
SA-MTL& + Multi-head (5) & 0.892\\
SA-MTL& + Position encoding & 0.892\\

\hline
\end{tabular}
\begin{tablenotes}\footnotesize
 \item[\dag] Self-attention module has a fully connected layer inside. The Hidden Unit Size 2 is used at the fully connected layer.
 \item[*] We experimented by replacing the first CNN layer of our model with an RNN layer.
 \item[**] We experimented by replacing the discrete output layer of our model with a max pooling layer.
\end{tablenotes}
\end{threeparttable}
\end{table}

\section{Conclusion}
Our SA-MTL model achieved better performance than the existing models on several datasets. Our model is descriptor-free as SMILES is the direct input of our model. We implemented the self-attention with the multi-task learning method to predict the chemical compound properties. We showed that the encoder part of the Transformer could serve as a classification predictor. Our SA-MTL model exhibited the state-of-the-art performance in the Tox21 and several other datasets.


\section*{Acknowledgment}
The study is supported by National Research Council of Science \& Technology (NST) grant by the Korea government (MSIP) (No. CAP-17- 01-KIST Europe).




%

\end{document}